\documentclass{article}
\usepackage[a4paper, total={6in, 8in}]{geometry}

\usepackage[utf8]{inputenc} % allow utf-8 input
\usepackage[T1]{fontenc}    % use 8-bit T1 fonts
\usepackage{hyperref}       % hyperlinks
\usepackage{url}            % simple URL typesetting
\usepackage{booktabs}       % professional-quality tables
\usepackage{amsfonts}       % blackboard math symbols
\usepackage{nicefrac}       % compact symbols for 1/2, etc.
\usepackage{microtype}      % microtypography
\usepackage{lipsum}
\usepackage{amsmath}
\usepackage{graphicx}
\usepackage{xcolor}
\usepackage{subcaption}
\usepackage{epstopdf}
\usepackage{xifthen}

\usepackage{cleveref} % load it last

\newcommand{\expected}[1]{\mathbb{E}\{#1\}}
\newcommand{\expectedv}[2]{\mathbb{E}_{#2}\{#1\}}
\newcommand{\grad}[1]{\delta^{#1}}
\newcommand{\actd}[1]{\phi'\left(#1\right)}
\newcommand{\act}[1]{\phi\left(#1\right)}
\newcommand{\der}[2]{\frac{\delta#1}{\delta#2}}
\newcommand*{\defeq}{\stackrel{\text{def}}{=}}
\newcommand*{\diag}[1]{\textrm{diag}\left(#1\right)}
\newcommand*{\loss}{\mathcal{L}}
\newcommand{\var}[1]{\sigma_{#1}^2}
\newcommand{\varw}[1]{\var{w}}
\newcommand*{\NTK}{\Theta}
\newcommand{\wvec}{w}
\newcommand{\Amap}[1][]{%
\ifthenelse{\isempty{#1}}{\mathcal{A}}{\mathcal{A}\left(#1\right)}%
}
\newcommand{\Vmap}[1][]{%
\ifthenelse{\isempty{#1}}{V}{V\left(#1\right)}%
}
\newcommand{\Vpmap}[1][]{%
\ifthenelse{\isempty{#1}}{V'}{V'\left(#1\right)}%
}
\newcommand*{\ch}{c}
\newcommand*{\corr}[1]{\textcolor{black}{#1}}
\newcommand*{\SigmaA}[1]{\Sigma_{{a}^{#1}}}
\newcommand*{\meanA}[1]{\mu_{{a}^{#1}}}
\newcommand*{\SigmaD}[1]{\Sigma_{\grad{#1}}}
\newcommand*{\SigmaZ}{\Sigma_{z}}
\newcommand*{\lr}{\eta}

\title{The Neural Tangent Link Between CNN Denoisers and Non-Local Filters}

\author{
  Juli\'an Tachella\thanks{The codes associated with this work are available at \href{https://gitlab.com/Tachella/neural_tangent_denoiser}{https://gitlab.com/Tachella/neural\_tangent\_denoiser}} \\
 School of Engineering\\
  University of Edinburgh\\
  \and
 Junqi Tang \\
 School of Engineering\\
  University of Edinburgh\\
  \and
 Michael E. Davies \\
 School of Engineering\\
  University of Edinburgh\\
}

\begin{document}
\maketitle

\begin{abstract}
Convolutional Neural Networks (CNNs) are now a well-established tool for solving computational imaging problems. Modern CNN-based algorithms obtain state-of-the-art performance in diverse image restoration problems. Furthermore, it has been recently shown that, despite being highly overparameterized, networks trained with a single corrupted image can still perform as well as fully trained networks. \corr{We introduce a formal link between such networks through their neural tangent kernel (NTK), and well-known non-local filtering techniques, such as non-local means or BM3D.} The filtering function associated with a given network architecture can be obtained in closed form without need to train the network, being fully characterized by the random initialization of the network weights. While the NTK theory accurately predicts the filter associated with networks trained using standard gradient descent, our analysis shows that it falls short to explain the behaviour of networks trained using the popular Adam optimizer. The latter achieves a larger change of weights in hidden layers, adapting the non-local filtering function during training. We evaluate our findings via extensive image denoising experiments.
\end{abstract}

\section{Introduction}
Convolutional neural networks are now ubiquitous in deep learning solutions for computational imaging and computer vision, ranging from image restoration tasks such as denoising, deblurring, inpainting and super-resolution, to image reconstruction tasks such as computed tomography~\cite{mardani2018neural} and magnetic resonance imaging~\cite{liu2020rare}. However, the empirical success of CNNs is in stark contrast with our theoretical understanding. Contrary to traditional sparse models~\cite{elad2010sparse}, there is little understanding of the implicit assumptions on the set of plausible signals imposed by  CNNs.

%A standard approach consists of training the networks with a large dataset of clean images, which may be not be available for many important applications such as dynamic MRI~\cite{liu2020rare}. Hence, a line of research has been devoted to reduce (or remove completely) the need for clean training images~\cite{lehtinen2018n2n,krull2019n2v,batson2019n2s}. 

Perhaps surprisingly, Ulyanov et al.~\cite{ulyanov2018deep} discovered that training a CNN only with a single corrupted image (the one being restored) could still achieve competitive reconstructions in comparison to fully trained networks, naming this phenomenon the deep image prior (DIP). This discovery challenges traditional wisdom that networks should be trained with large amounts of data and illustrates the powerful bias of CNN architectures towards natural images. \corr{Similar ideas have} also been explored in Noise2Self~\cite{batson2019n2s} and other variants~\cite{krull2019n2v}. In this setting, the number of weights (e.g., 2,000,000 for a U-Net CNN~\cite{ulyanov2018deep,ronneberger2015unet}) is much larger than the number of pixels in the training image (e.g., 50,000 pixels of a standard $128\times128$ color image). The clean version of the corrupted image is obtained by early-stopping the optimization process before the network fully matches the noisy image or by considering a loss that does not allow the network to learn the corrupted image exactly~\cite{batson2019n2s}. These surprising results raise the following questions: how, amongst all possible optimization trajectories towards the multiple global minima of the training loss, the procedure consistently provides close to state-of-the-art reconstructions? What is the role of the optimization algorithm on the trajectory towards the global minima, and how does it affect the bias towards clean images?

\begin{figure*}
   \centering
     \includegraphics[width=1\textwidth]{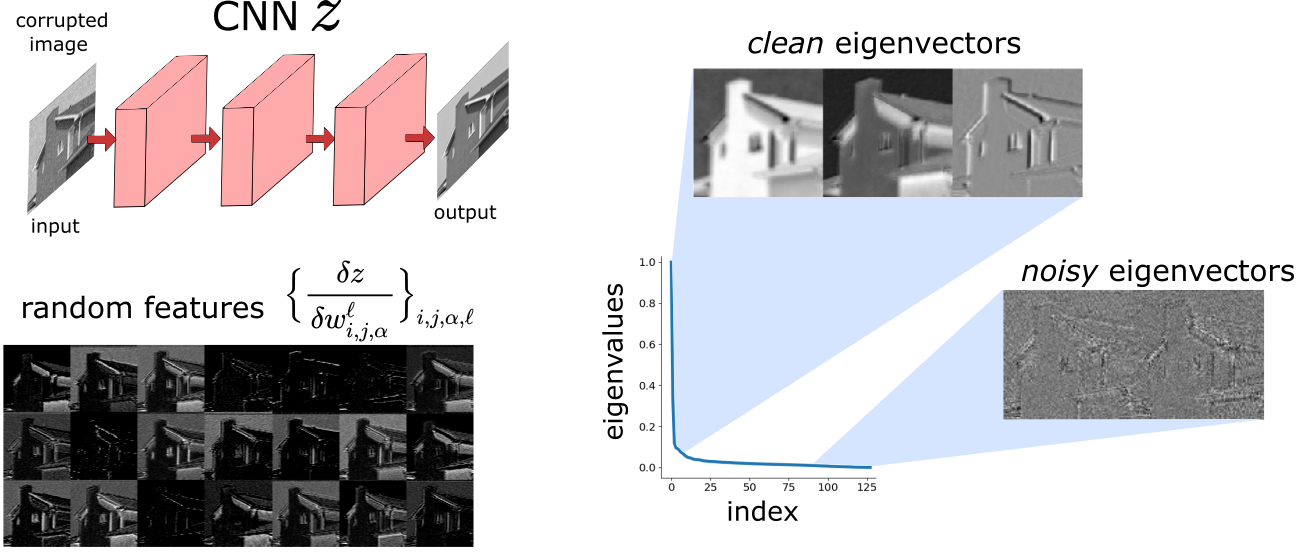}
   \caption{A convolutional neural network $z$ trained with gradient descent on a single corrupted image can achieve powerful denoising. The left eigenvectors of the Jacobian provide a representation based on patch similarities which is robust to noise.}
   \label{fig:schematic_ntd}
\end{figure*}

Despite their surprisingly good performance, these methods provide comparable or slightly worse denoising results than classical patch-based non-local filtering techniques, such as non-local means (NLM)~\cite{buades2005non} or BM3D~\cite{dabov2007image}, which also only have access to the corrupted image. Moreover, training a large neural network is more computationally intensive. Subsequent questions then arise: is the neural network performing a similar filtering process? Can we avoid the slow training, and apply this filter in a more direct way? \corr{These insights are important to build a better framework in which we can optimize and design new denoisers and other low-level computer vision algorithms.}

Denoising is generally considered as the fundamental building block of any image restoration problem. In many applications, CNNs are used to perform denoising steps, either in unrolled schemes~\cite{mardani2018neural} or in the context of plug-and-play methods~\cite{romano2017red,venkatakrishnan2013plug}. Hence, understanding better the bias of CNNs towards clean images is the first step towards more general imaging problems.

On another line of work, researchers have also observed that increasing the amount of overparameterization does not necessarily harm the generalization of the network~\cite{zhang2016understanding} in the context of classification. Recently, Jacot et al. showed that overparameterized neural networks trained with (stochastic) gradient descent (GD) converge to a Gaussian process as the number of weights tends to infinity, with a kernel that depends only on the architecture and variance of the random initialization, named the neural tangent kernel (NTK)~\cite{jacot2018ntk}. While the properties and accuracy of the kernel were analyzed for image classification~\cite{arora2019cntk}, to the best of our knowledge, little is known in the context of high-dimensional image restoration with no clean data. Can this theory explain the good denoising performance of networks trained with a single corrupted image?

In this paper, we study overparameterized convolutional networks and their associated neural tangent kernel in the context of the image denoising, formalizing  strong links with classical non-local filtering techniques, but also analyzing the short-comings of this theory to fully explain the results obtained by the DIP.
The main contributions of this paper are as follows:
\begin{enumerate}
    \item We show that GD trained CNN denoisers with a single corrupted image (placed both at the input and as a target) in the overparameterized regime equate to performing an existing iterative non-local filtering technique known as \emph{twicing}~\cite{milanfar2012tour}, where the non-local filter is characterized by the architectural properties of the network.   Moreover, \corr{these filters impose a form of low-dimensionality due to their fast eigenvalue decay}, and efficient filtering can be performed directly without the CNN, using the Nystr\"{o}m approximation~\cite{seeger2001nystrom}.
    \item Departing from previous explanations~\cite{cheng2019bdip,Soltanolkotabi2020Denoising}, we show that the DIP cannot be solely understood as a prior promoting low-pass images. We link this short-coming to the choice of the optimization algorithm.  When trained with GD, the DIP has poor performance as predicted by the NTK theory, and maintains a fixed low-pass filter throughout training. However, training with the popular Adam optimizer as in the original DIP is able to adapt the filter with non-local information from the target image.
    \item We evaluate our findings with a series of denoising experiments, showing that the fixed non-local filter associated with gradient descent performs significantly better when the corrupted image is placed at the input, whereas the Adam optimizer adapts the filter during training, providing good results for both scenarios.
\end{enumerate}

\begin{figure*}
   \centering
     \includegraphics[width=.9\textwidth]{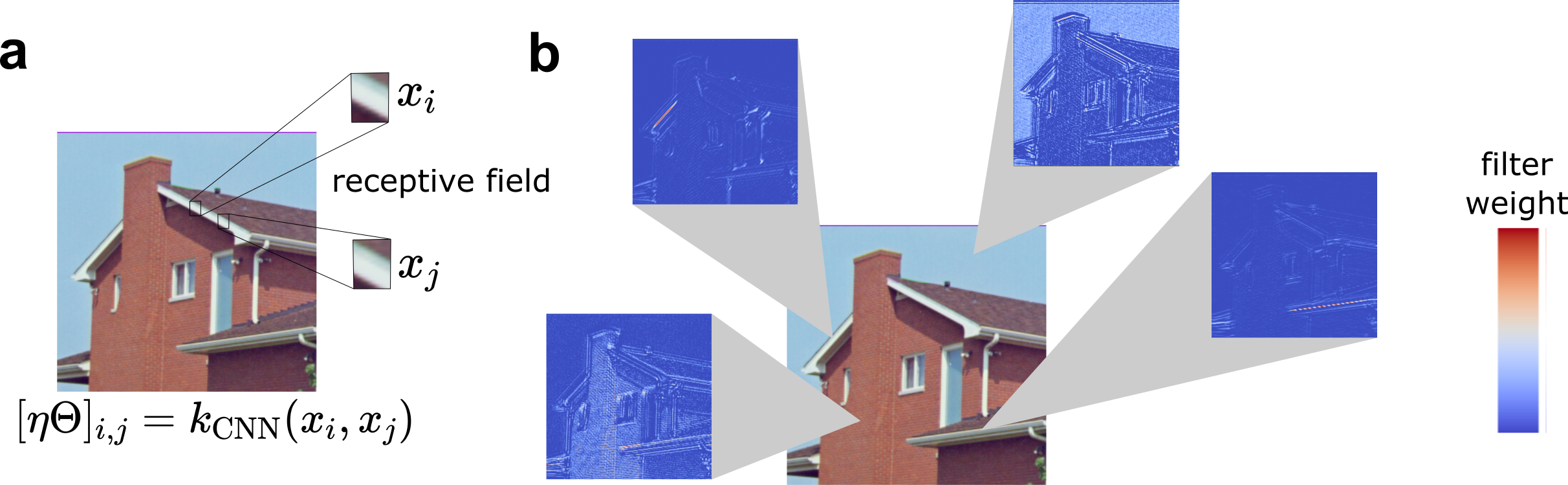}
   \caption{Non-local filter associated to the tangent kernel of a CNN with a single hidden layer. a) The filter can be obtained in closed form as the number of channels tends to infinity, where each $(i,j)$th entry corresponds to the similarity between the patches centered at pixels $i$ and $j$. b) Filter weights for different pixels in the house image, where red/white indicates a higher weight, and blue indicates a zero weight.}
   \label{fig:example_ntd}
\end{figure*}

% The paper is organized as follows: \Cref{sec:related work} enumerates previous works related to our findings. \Cref{sec:preliminaries} introduces some basic concepts and notation. \Cref{sec:twicing} presents the filtering function associated with overparameterized CNNs trained with gradient descent. The adaptation related to the Adam optimizer is discussed in \Cref{sec:adaptive}. Finally, \Cref{sec:experiments} illustrates the differences with several examples. 
\section{Related Work} \label{sec:related work}
\textbf{Neural networks as Gaussian processes:}
Neal~\cite{neal1995bayesian} showed that a randomly initialized fully-connected networks converge to Gaussian process. This result was recently extended to the convolutional case~\cite{novak2019bayesian}. Jacot et al.~\cite{jacot2018ntk} showed that the network remains a Gaussian process throughout GD training, but with a different kernel, the NTK. Arora et al.~\cite{arora2019cntk} studied the kernel of a convolutional architecture for image classification, while Yang~\cite{yang2019mean} extended these results to a wider set of architectures. All these works focus on classification with a set of training pairs of images and labels, whereas we study high-dimensional regression (denoising) with no clean training data.

\textbf{Non-local (global) filtering:}
A powerful class of denoisers in image processing use patch-based filtering, e.g., NLM~\cite{buades2005non} and BM3D~\cite{dabov2007image}. Milanfar studied these from a kernel function perspective~\cite{milanfar2012tour}, identifying the associated affinity (kernel) matrices, along with different iterative denoising techniques.

\textbf{CNNs as non-local filters:}
Recently, Mohan et al.~\cite{Mohan2020Robust} showed that a fully-trained denoising CNN without biases can be interpreted as a non-local filter by examining the input-output Jacobian of the network. They perform a local analysis of trained networks, whereas we study the global convergence during training, providing analytical expressions for the filters.

\textbf{Self-supervised image denoising:}
In Noise2Noise~\cite{lehtinen2018n2n}, the authors show that training a denoising network with noisy targets can achieve similar performance to a network trained with clean targets. Noise2Void~\cite{krull2019n2v} and Noise2Self~\cite{batson2019n2s} present a self-supervised training procedure that achieves good performance, even with a single noisy image.

\textbf{Deep image prior interpretations:}
Cheng et al.~\cite{cheng2019bdip}, analyzed the spatial (low-pass) filter associated to a U-Net CNN at initialization, following the Gaussian process interpretation of~\cite{neal1995bayesian}. Similarly, Heckel and Soltanolkotabi~\cite{Soltanolkotabi2020Denoising} show that CNN decoders generate low-pass filters under GD learning, and attribute the DIP's success to this. Our work differs significantly from theirs, as we study the non-local filter behaviour of the learning process, showing that the low-pass filter behaviour does not explain DIP's state-of-the-art performance. In contrast to the spatial filters in~\cite{cheng2019bdip,Soltanolkotabi2020Denoising}, the induced filters studied here can be made dependent on non-local information of the corrupted image, hence providing competitive performance to other patch-based methods.

\section{Preliminaries} \label{sec:preliminaries}
\subsection{Convolutional neural networks}
An $L$-layer vanilla\footnote{While our derivations focus on a simple CNN structure for the sake of clarity of the presentation, the analysis can be extended to account for multiple channels at the input and output (e.g., RGB images), biases, skip connections, downsampling and upsampling operations, see Appendices A, E and F.} convolutional neural network with $\ch$ channels at each hidden layer is defined as
\begin{align}
    a^{1}_{i}(x) &= W_{i,1}^{1}x 
\end{align}
\begin{align} \label{eq:cnn}
    a^{\ell}_{i}(x) &= \sum_{j=1}^{\ch} W_{i,j}^{\ell}\act{a_{j}^{\ell-1}(x)} \\
    z(x) &= \sum_{j=1}^{\ch} W_{1,j}^{L}\act{a_{j}^{L-1}(x)}
\end{align}
where $\phi:\mathbb{R}^{d}\mapsto\mathbb{R}^{d}$ is an element-wise activation function, $a^{\ell}_{i}\in\mathbb{R}^{d}$ are the $i$th channel preactivations at layer $\ell$, $W_{i,j}^{\ell}\in \mathbb{R}^{d\times d}$ are circulant matrices associated with convolution kernels of size $r\times r$ with trainable parameters $\{ w_{i,j,\alpha}^{\ell}: \alpha=1,\dots,r^2 \}$, the input and output are vectorized images of $d$ pixels, denoted as $x\in \mathbb{R}^{d}$ and $z\in \mathbb{R}^{d}$ respectively.
We focus on restoration with no clean training data, where only the corrupted image $y$ is available as a training target. For the input there are 2  options:
\begin{enumerate}
    \item Corrupted image: we place the noisy target as the input, i.e., $x=y$, i.e., Noise2Self and variants~\cite{krull2019n2v,batson2019n2s}.
    \item Noise: the input is assigned with iid noise, i.e., $x\sim\mathcal{N}(0,I)$, i.e., the DIP setting~\cite{ulyanov2018deep}.
\end{enumerate}
As there is a single input to the network, we will drop the dependence of $z$ with respect to $x$ for the sake of clarity, only focusing on the dependence with respect to the weights, denoted as $z(w)$, where the high-dimensional vector $\wvec$ contains all individual weights $w_{i,j,\alpha}^{\ell}$. We assume that the weights of the network are drawn iid using the standard \emph{He initialization}~\cite{He2015delving}, $w_{i,1,\alpha}^{1}\sim \mathcal{N}(0,\frac{\sigma_w^{2}}{r^2})$ and $w_{i,j,\alpha}^{\ell}\sim \mathcal{N}(0,\frac{\sigma_w^{2}}{r^2\ch})$ for $\ell=2,\dots,L$, which avoids vanishing or divergent outputs in deep networks, where $\sigma_w^{2}$ is chosen depending on the non-linearity~\cite{xiao2018dynamical}, e.g., $\sigma_w^{2}=2$ for relu.  As in most image restoration problems, we assume training is performed on the squared loss, defined as $\loss(w) = \frac{1}{2} ||z(\wvec)-y||_2^2$. %The DIP minimizes the loss using Adam~\cite{kingma2015adam}, early-stopping before the network overfits the corrupted image $y$.

\subsection{Non-local denoisers}
Multiple existing non-local collaborative filtering techniques~\cite{milanfar2012tour}, such as the well-known NLM, BM3D or LARK~\cite{takeda2007lark}, consist in computing a filtering matrix $W$ with the $(i,j)$th entry related to the affinity between a (noisy) image patch \corr{$y_i$} centered at pixel $i$ and another (noisy) image patch \corr{$y_j$} centered at pixel $j$. 
For example, the NLM affinity function\footnote{There is a subtle, but important point: the NLM filter matrix is normalized~\cite{milanfar2012tour} as $W' = \text{diag}(1/1^TW)W$ or using Sinkhorn's positive semidefinite approximation.} with patch size of $r\times r$ and parameter $\sigma^2$ is \corr{
\begin{equation}
   [W]_{i,j} =  k_{\text{NLM}}(y_i,y_j) = e^{-\frac{||y_i-y_j||_2^2}{\sigma^2}}
\end{equation}}
The most basic denoising procedure\footnote{Although this procedure seems to be linear it is in fact nonlinear due to the dependence of $W$ on $y$.} consists of applying $W$ directly to the noisy image $z=Wy$. However, the performance can be improved using an iterative procedure named \emph{twicing}~\cite{milanfar2012tour}
\begin{equation}
\label{eq:twicing}
    z^{t+1} = z^{t} + W(y-z^{t})
\end{equation}
Given a fixed positive semidefinite filter matrix with eigendecomposition $W=V\diag{\lambda_1,\dots,\lambda_d}V^T$, we can express the output in the orthogonal basis $V$, i.e., 
\begin{equation}
    z^{t} = \sum_{i=1}^{d} (1-(1-\lambda_i)^{t}) (v_i^{T}y)  v_i 
\end{equation}
where $v_i$ is the $i$th column of $V$ and $0\leq\lambda_i<2$.
 Assuming that $W$ is approximately independent of the noise~\cite{milanfar2012tour}, the mean squared error (MSE) can be easily estimated as 
\begin{equation}
\label{eq:bias_variance}
     \text{MSE} \approx \sum_{i=1}^{d} (1-\lambda_i)^{2t} (v_i^T\hat{x})^2 +(1-(1-\lambda_i)^{t})^{2} \sigma^2 
\end{equation}
where $\hat{x}$ denotes the noiseless image, and the first and second terms represent the \corr{(squared)} bias and variance respectively. 
As it can be seen in~\cref{eq:bias_variance}, the twicing strategy trades bias for variance, starting with a blurry estimate and converging towards the noisy target $y$ as $t\to\infty$. As with the early-stopped neural networks, the procedure is stopped before overfitting the noise. 

For a fixed signal-to-noise ratio, the denoising performance will depend on how concentrated is the energy of the signal $x$ on the leading eigenvectors of $V$ (controlling the bias term) and how fast is the decay of the eigenvalues of the filter (controlling the variance term). It will also depend on how close are the computed non-local similarities using the noisy image from the oracle ones (computed with the clean image). For example, BM3D also adapts the filtering matrix, by using a prefiltered version of the noisy image to calculate the affinity between pixels~\cite{dabov2007image}. 

\section{\corr{Neural tangent kernel analysis}}\label{sec:twicing}
The seminal work in~\cite{jacot2018ntk}, and subsequent works~\cite{lee2019wide,yang2019mean,arora2019cntk}, pointed out that as the number of parameters goes to infinity, which equates to taking $\ch\to\infty$, a network trained with GD and learning rate $\eta$ of order\footnote{The learning rate cannot be larger than $\mathcal{O}(\ch^{-1})$ in order to converge to a global minimum~\cite{karakida2019fim}. We have also observed in our experiments using the larger learning rates leads to a divergent output.} $\mathcal{O}(\ch^{-1})$, leads to a vanishingly small change of each individual weight~\cite{lee2019wide, arora2019cntk}
\begin{equation}
\label{eq:weight_change_gd}
   \max_{t} |(w_{i,j,\alpha}^{\ell})^{t}-(w_{i,j,\alpha}^{\ell})^{0}| = \begin{cases} 
     \mathcal{O}(\ch^{-1}) & \mbox{if } \ell=L \\
    \mathcal{O}(\ch^{-3/2}) & \mbox{otherwise} \end{cases}
\end{equation}
where $t$ denotes the gradient descent iteration, such that the overall change of the parameter vector $||\wvec-\wvec^{0}||_2$ is of order $\mathcal{O}(\ch^{-1/2})$. Hence, the evolution of the network can be well described by a  first order expansion around the random initialization, $z(\wvec^{t}) \approx z(\wvec^{0})+ \der{z}{\wvec} (\wvec^{t}-\wvec^{0})$, where $\der{z}{\wvec}$ is the Jacobian of the network at initialization, whose columns are shown in \Cref{fig:schematic_ntd}. In this regime, the training dynamics reduce to
\begin{equation}\label{eq:twicing_ntk}
    z^{t+1} = z^t + \lr\NTK_{L}^{0} (y-z^{t}) 
\end{equation}
with $z^0=z(\wvec^{0})$ and the positive semidefinite NTK Gram matrix (1 training sample and $d$ outputs) given by
\begin{align}
    \NTK_{L}^{0} &= \left. \der{z}{\wvec} (\der{z}{\wvec})^T\right|_{\wvec=\wvec^{0}} 
    %\\ &= \sum_{\ell,i,j,\alpha}   \der{z}{w_{i,j,\alpha}^{\ell}} (\der{z}{w_{i,j,\alpha}^{\ell}})^T 
\end{align}
which stays constant throughout training as $\ch\to\infty$.

The denoising process in \cref{eq:twicing_ntk} is identical to the twicing procedure in \cref{eq:twicing}, where the \corr{filter $W$ is given by the non-local affinity matrix} $\lr\NTK_L$.
The resulting pixel affinity function depends on the architecture, such as depth, convolution kernel size and choice of non-linearity. The size of each patch is given by the network's receptive field, as illustrated in \Cref{fig:example_ntd}.  As with non-local filters, the denoising performance depends on the alignment between the noiseless image and the leading eigenvectors \corr{of $\lr\NTK_L$}. As shown in \Cref{fig:schematic_ntd}, the filter associated with a CNN exhibits a fast decay of its eigenvalues and the image contains most of its energy within the leading eigenvectors. 

The filter $\lr\NTK$ can be computed in closed form via the following recursion\footnote{A detailed derivation is provided in Appendix D.}~\cite{arora2019cntk} 
\begin{equation}
\label{eq:recursive_ntk}
   \begin{cases}
    \SigmaA{\ell} = \Amap[\Vmap[\SigmaA{\ell-1}]] \\
    \lr\NTK_{\ell} = \SigmaA{\ell} + \Amap[\Vpmap[\Sigma_{a^{\ell-1}}]\circ\lr\NTK_{\ell-1}]
   \end{cases}
\end{equation} 
with base case (one hidden layer)
\begin{equation}
 \SigmaA{2} = \eta\NTK_{2} = \Vmap[\Amap[xx^T]]
\end{equation}
where $\circ$ denotes element-wise matrix multiplication, and $\SigmaA{\ell}$ denotes the covariance of the preactivations of the network $a^{\ell}_i$ for all $i=1,\dots,\ch$.  The convolution operator  related to a filter size of $r\times r$ pixels is a mapping between positive semidefinite matrices $\Amap[]: \text{PSD}_d \mapsto \text{PSD}_d$ defined as~\cite{xiao2018dynamical} 
\begin{equation}
    [\Amap[\Sigma]]_{i,j} =  \frac{1}{r^2}\sum_{i', j'} [\Sigma]_{i', j'}
\end{equation}
where $i'$ and $j'$ indicate the pixels within patches of size $r \times r$ centered at pixels $i$ and $j$ respectively.
The maps $\Vmap[]: \text{PSD}_d \mapsto \text{PSD}_d$ and $\Vpmap[]: \text{PSD}_d \mapsto \text{PSD}_d$  are defined by the choice of non-linearity and its derivative as 
\begin{align}
    \Vmap[\Sigma] &= \sigma_w^{2}\expectedv{\act{h}\act{h^T}}{h\sim \mathcal{N}(0,\Sigma)} \\
     \Vpmap[\Sigma] &= \sigma_w^{2}\expectedv{\actd{h}\actd{h^T}}{h\sim \mathcal{N}(0,\Sigma)}
\end{align}
which are available in closed form for many popular non-linearities including relu (see Appendix B).

For example, a relu CNN with a single hidden layer and a convolution kernel size of $r \times r$ pixels has an associated affinity function
\begin{equation}
\label{eq:single_layer_ntd}
     k_{\text{CNN}}(x_i,x_j) = \frac{||x_i||_2 ||x_j||_2}{\pi} (\sin(\varphi) + (\pi-\varphi)\cos(\varphi))
\end{equation}
with $\varphi = \arccos \frac{x_i^Tx_j}{||x_i||_2 ||x_j||_2}$. \Cref{fig:example_ntd} illustrates the filter computed using the closed-form kernel in \cref{eq:single_layer_ntd}, \corr{which weights similar patches more strongly.}

\subsection{Computing the analytic filter}
Instead of training a neural network as in the DIP, we can explicitly compute the filtering matrix $\lr\NTK$, and use \cref{eq:twicing} to perform the denoising. As the size of the filter matrix ($d\times d$) is prohibitively big to compute and store for large images, we instead only compute a random selection of $m \ll d$ columns of $\lr\NTK$, and approximate the matrix with its leading eigenvalues and eigenimages using the Nystr\"{o}m method~\cite{seeger2001nystrom}. The columns are chosen by selecting random pixels uniformly distributed in space, as in the global image denoising algorithm~\cite{talebi2014glide}. A detailed description of the algorithm can be found in Appendix G.

\begin{figure*}
  \centering
    \includegraphics[width=1\textwidth]{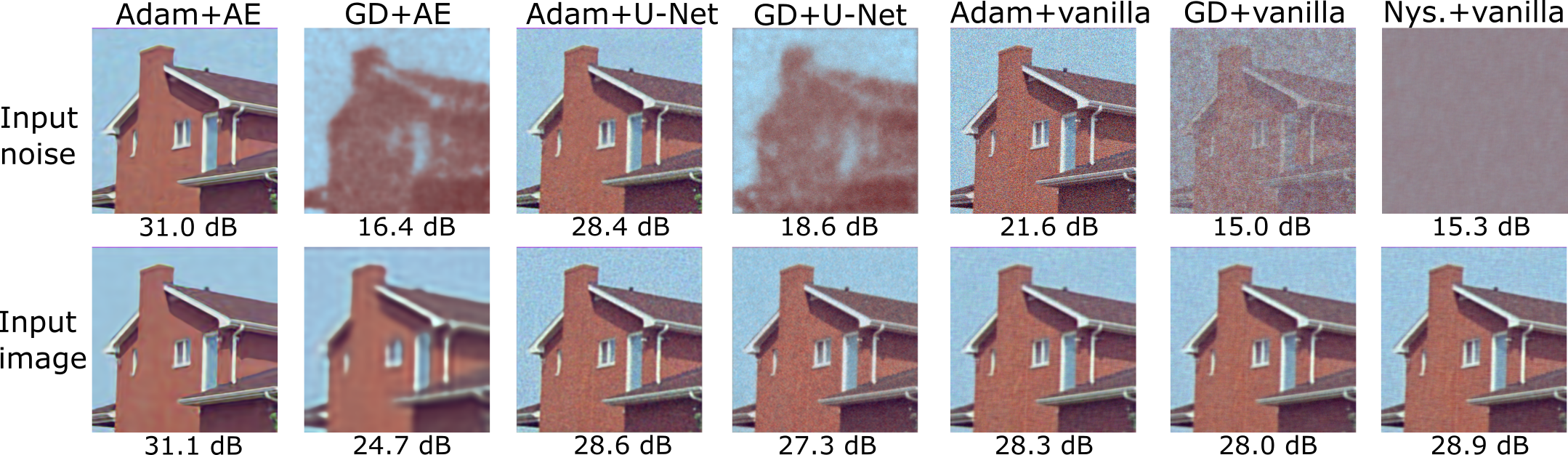}
  \caption{Results for the `house' image. PSNR values are reported below each restored image. The best results are obtained by the autoencoder trained with Adam, which is able to provide smoother estimates while preserving sharp edges. However, it provides worse estimates of images with noise-like textures, such as the `baboon' image (see Appendix I).}
  \label{fig:house comparison}
\end{figure*}

\section{Adaptive filtering and the deep image prior} \label{sec:adaptive}
\subsection{The DIP is not a low-pass filter} \label{subsec:ntk_failing}
In the DIP paper, the input is chosen to be random iid noise. In this case, the resulting filter $\lr\NTK$ does not depend in any way on the target image $y$, \corr{and} the non-local similarities are computed using the input noise. How bad can this filter be? Applying \cref{eq:recursive_ntk} with noise at the input we get in expectation
\begin{equation}
\label{eq:cov_gp}
    [\lr\NTK]_{i,j} = \frac{1}{d} \begin{cases} 
   1 &\mbox{if } i=j \\
    \kappa_L & \mbox{otherwise} \end{cases}
\end{equation}
with $\kappa_L\approx 0.25$ for large $L$, which has a very large first eigenvalue $\lambda_1=(1-\kappa_{L})/d+ \kappa_L\approx 0.25$ associated with a constant image $v_1=[1,\dots,1]^T/\sqrt{d}$ and the rest of the eigenvalues of small size $\lambda_{i}=0.75/d$ for $i=2,\dots,d$. Hence, this (linear) filter would just be useful for constant images. In the case of an autoencoder (AE) architecture with downsampling and upsampling layers, the resulting filter is a crude low-pass filter, but still does not depend on the target image. Previous works~\cite{cheng2019bdip,Soltanolkotabi2020Denoising} hypothesized that this filter can explain the bias towards clean images. However, it is well-known that low-pass filters don't provide good denoising results, as they tend to oversmooth the edges and fine details of the image. We show that this gap between theory and practice is because the DIP in~\cite{ulyanov2018deep} is not trained with GD but Adam.

\begin{table*}[!h]
\centering
\begin{tabular}{l|c|c|c|c|c|c|}
\cline{2-7}
 & \multicolumn{2}{c|}{Vanilla CNN} & \multicolumn{2}{c|}{U-Net}  & \multicolumn{2}{c|}{Autoencoder} \\ 
\cline{2-7}
 & Noise & Image & Noise & Image & Noise & Image \\ \hline
\multicolumn{1}{|l|}{Adam} & 19.6 & 27.4 &  28.3 & 28.1& 29.2 & \textbf{29.3} \\ \hline
\multicolumn{1}{|l|}{Gradient descent} &  15.2 & 27.5 & 16.5 & 27.1  & 15.0 & 26.8 \\ \hline
\multicolumn{1}{|l|}{Nystr\"{o}m} & 15.2 & 28.3  \\ \cline{1-3}
\end{tabular}
\caption{Average peak-signal-to-noise ratio (PSNR) [dB] achieved by different combinations of network architecture, input and optimizer on the dataset of 9 color images~\cite{dabov2007image}.}
\label{tab:psnr_results}
\end{table*}

\subsection{Adaptive filtering with Adam}
The Adam optimizer updates the weights according to
\begin{align}
    \wvec^{t+1} &= \wvec^{t} -\tilde{\lr} H^{t} \der{\loss}{\wvec}(w^{t}) + \beta_1(\wvec^{t} -\wvec^{t-1})
\end{align}
where $\beta_1$ is a hyperparameter controlling the momentum and learning rate,  $\tilde{\lr}= \lr(1-\beta_1)$  and $H^{t}$ is a diagonal matrix containing the inverse of a running average of the squared value of the gradients, computed using the hyperparameter $\beta_2$. The resulting filter is adapted at each iteration,
\begin{equation}
    \tilde{\NTK}_{L}^{t} = \left. \der{z}{\wvec} H^{t}(\der{z}{\wvec})^T\right|_{\wvec=\wvec^{t}} 
\end{equation}
and the denoising process can be written as 
\begin{equation}
\label{eq:output_dynamics}
    z^{t+1} = z^t + \tilde{\lr}\tilde{\NTK}_L^{t} (y-z^{t}) + \beta_1(z^{t} -z^{t-1})
\end{equation}
%where the filtering matrix $\tilde{\lr}\tilde{\NTK}_L^{t}$ adapts according to the residuals $(y-z^{t})$.
The matrix $H^{k}$ imposes a metric in the weight space which differs from the standard Euclidean metric of GD. 
Unfortunately, as shown by Gunasekar et al.~\cite{gunasekar2018implicit}, this metric  depends on the choice of the learning rate, rendering a general analysis of the adaptation intractable\footnote{Removing the adaptation ($\beta_1,\beta_2\to 0$) the algorithm reduces to sign gradient descent, i.e., steepest descent with respect to the $\ell_\infty$ norm~\cite{balles2017dissecting}, but it is still sensitive to the choice of learning rate~\cite{gunasekar2018implicit}}. Moreover, as shown in our experiments, all weights, including intermediate layers, undergo a larger change than in gradient descent, that is 
\begin{equation}
\label{eq:weight_change_adam}
   \max_t |(w_{i,j,\alpha}^{\ell})^{t}-(w_{i,j,\alpha}^{\ell})^{0}| = \mathcal{O}(\ch^{-1}) \quad \forall \ell=1,\dots,L
\end{equation}
such that the overall change of the parameter vector $||\wvec^{t}-\wvec^{0}||_2$ is $\mathcal{O}(1)$, and a Taylor expansion around the initialization does not model accurately the training dynamics\footnote{Note that a higher order expansion~\cite{bai2020taylorized} cannot explain the good performance of the DIP, as higher order derivatives are still independent of the target. Moreover, the Hessian would not describe $\mathcal{O}(1)$ perturbations.}. 
 Nonetheless, here we provide insight into how the resulting filtering kernel \corr{can still absorb non-local properties from the \emph{target output}}. Similarly to the output dynamics, the evolution of the preactivations can be well described by its \corr{(time-varying)} first order expansion\footnote{\corr{Here we assume $\beta_1=0$ for simplicity.}}:
\begin{align} \label{eq:ev_preact}
 (a^{\ell}_i)^{t+1} &\approx (a^\ell_i)^{t} - \lr \der{a^\ell_i}{\wvec} H^{k} (\der{a^\ell_i}{\wvec})^T \der{\loss}{a^\ell_i} 
 \\ 
 &\approx (a^\ell_i)^{t} - \lr \tilde{\NTK}_{\ell}^{t} (\grad{\ell}_i)^{t}
\end{align}
where the error gradient at layer $\ell$ and channel $i$ is defined as $\grad{\ell}_i\defeq \der{\loss}{a^\ell_i}\in \mathbb{R}^{d}$. 
%As the number of channels goes to infinity, if we treat the $\grad{\ell}_i$ as independent of the pre-activations $a_i^{\ell}$ computed in the forward pass~\cite{xiao2018dynamical,yang2019mean}, the gradients of a given layer can still be described by its second order moments. 
At initialization, this vector carries non-local information (via operator $\Amap[]$) about the target $y$, with covariance given by the recursion~\cite{xiao2018dynamical,yang2019mean}
\begin{align}\label{eq:sigmaD recursive}
    \SigmaD{\ell}  &=  \Amap[\SigmaD{\ell+1}] \circ \Vpmap[\SigmaA{\ell}]
\end{align}
starting with 
\begin{equation}\label{eq:sigmaD base}
\SigmaD{L-1} = \ch^{-1} \Amap[(y-z^{t})(y-z^{t})^T] \circ \Vpmap[\SigmaA{L-1}]
\end{equation}
which depends on the target image via the residual $(y-z^{t})$. A full derivation of \cref{eq:sigmaD recursive} is provided in Appendix C.
In the case of GD training, the change in the preactivations from initialization is negligible as the error terms $\grad{\ell}_i$ are of order $\mathcal{O}(\ch^{-1/2})$ due to the $\ch^{-1}$ scaling in \cref{eq:sigmaD base}. However, the larger change in intermediate layers when using Adam yields a non-negligible change in the preactivations, while the exact adaptation depends on the choice of hyperparameters $\lr$, $\beta_1$ and $\beta_2$. \corr{This larger change lies at the heart of the improved performance of the DIP in comparison with GD training.}

\begin{figure*}
    \centering
    \includegraphics[width=1\textwidth]{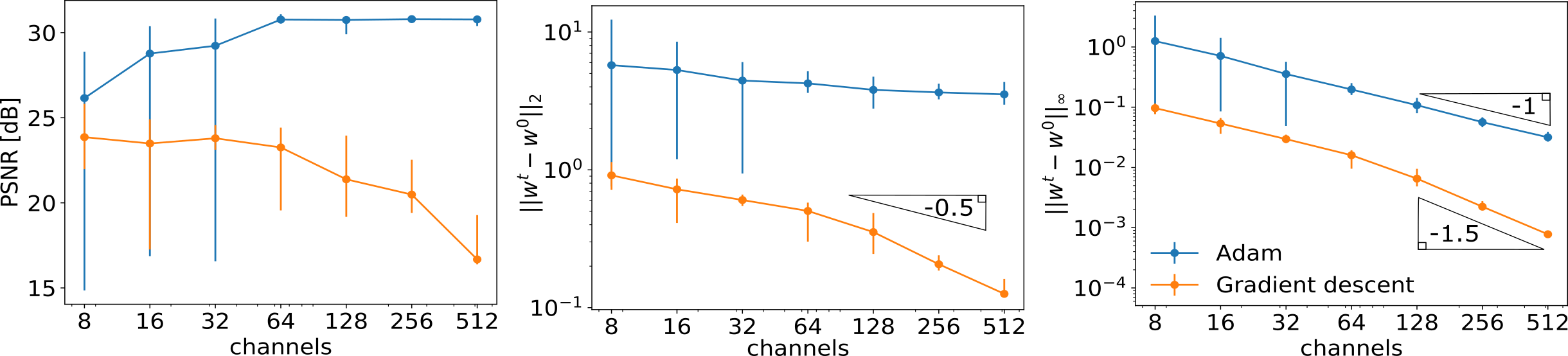}
    \caption{Comparison of Adam and GD training of an autoencoder with noise at the input as a function of the number of channels. The PSNR for the `house' image is shown on the left plot, whereas the average $\ell_2$ and $\ell_\infty$ change of weights in hidden layers is shown on the center and right plots respectively. The error bars denote the maximum and minimum values obtained in 10 Monte Carlo repetitions.}
    \label{fig:weight_change}
\end{figure*}

\section{Experiments} \label{sec:experiments}

We analyze the performance of \corr{single-image} denoising neural networks both with the corrupted image or iid noise at the input of the network on a standard dataset of 9 color images~\cite{dabov2007image} corrupted with Gaussian noise of $\sigma=25$. We evaluate 3 different architectures, a simple vanilla CNN with a single hidden layer and a kernel size of $11\times11$ pixels, a U-Net with 3 downsampling and upsampling stages and a kernel size of $3\times3$ pixels and an autoencoder with the same architecture as the U-Net but no skip-connections. All architectures use relu non-linearities. A detailed description of the chosen architectures can be found in Appendix H. For each combination of input and architecture, we optimize the network using Adam with standard hyperparameters ($\beta_1=0.9$ and $\beta_2=0.99$) and vanilla GD (no momentum). We also include results achieved by taking the infinite channel limit of the vanilla CNN, and computing the associated NTD filter. In this case, we use the Nystr\"{o}m approximation to reduce the memory requirements of storing the full matrix $\eta\NTK$. We found that computing only 2\% of its columns gives a negligible reduction of performance with respect to computing the full matrix. In the experiments with the image at the input, we remove the random initial output by redefining the network function as $\tilde{z}=z-z^{0}$ with a fixed translation $z^{0}$, such that $\tilde{z}^0 = 0$ (as with standard twicing). We run the optimization until there is no further improvement of the peak \corr{signal-to-noise ratio} (PSNR)\footnote{While we use the oracle image for a fair comparison of all methods, a SURE estimator of the mean squared error~\cite{ramani2008sure} could be used in practical applications.} or a maximum of $10^{6}$ iterations is reached, and keep the best performing output.

 \begin{figure*}
   \centering
     \includegraphics[width=1\textwidth]{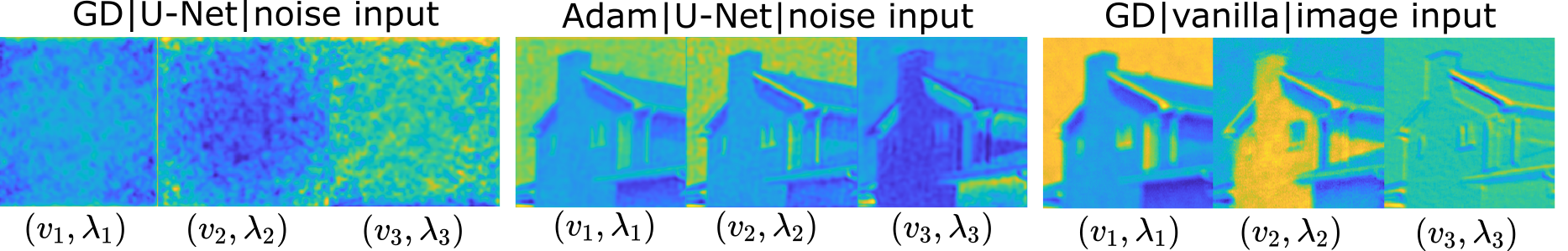}
   \caption{First 3 leading eigenvectors of the covariance of the last preactivations, $\SigmaA{L-1}$, after 500 iterations of training with Adam or gradient descent with different inputs (noise or image).}
   \label{fig:preact}
\end{figure*}

\subsection{Denoising performance}
The average PSNR obtained by all possible configurations is shown in \Cref{tab:psnr_results}. The results for one of the images in the dataset are shown in \Cref{fig:house comparison}. 

The best performances are achieved by the autoencoder architecture optimized with Adam, followed by the induced filter of the vanilla CNN, computed with the Nystr\"{o}m approximation. It is worth noting that while the autoencoder in the DIP uses batch normalization, biases, leaky relus instead of relus and a Swish activation function at the output, it does not perform significantly better without them (same average PSNR as the results reported in~\cite{ulyanov2018deep} and 0.1 dB improvement when placing the corrupted image at the input). Furthermore, the best results are obtained when placing the \corr{corrupted} image at the input, without requiring the carefully-designed loss functions of Noise2Void and Noise2Self. 

As predicted in \Cref{subsec:ntk_failing}, GD provides very poor reconstructions when inputting noise, but improves considerably with the corrupted image as the input\corr{, as only the latter has access to the non-local structure.} 
 While the vanilla CNN trained with GD and its Nystr\"{o}m approximation should in theory perform the same, the difference can be attributed to Nystr\"{o}m's lower rank approximation. Even though Adam plays a big role in adapting the autoencoder filter (8 hidden layers), it does not modify significantly the filter associated with a single hidden layer vanilla CNN. Denoising using the Nystr\"{o}m approximation of the analytic filter takes an average of 3 seconds per image, while training the autoencoder with Adam required 806 seconds\footnote{All the experiments were run with a GPU NVIDIA GTX 1080 Ti using the PyTorch library.}. This significant difference illustrates the potential speed up that can be obtained by having a better theoretical understanding of the denoising network.

\begin{figure}
   \centering
     \includegraphics[width=.6\columnwidth]{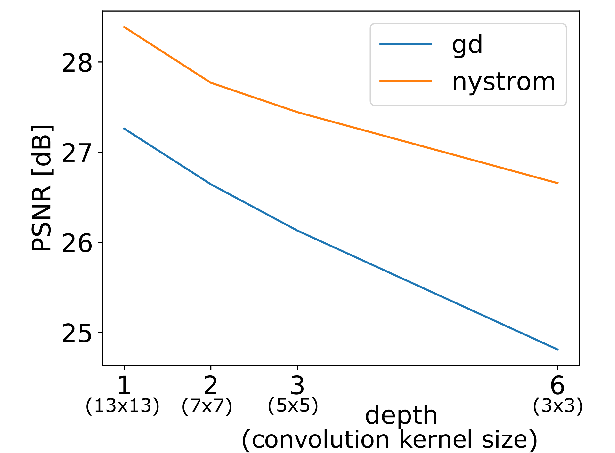}
   \caption{Denoising performance of the vanilla CNN on the dataset of 9 images for varying depth and a fixed total receptive field of $13\times13$ pixels.}
   \label{fig:depth}
\end{figure}

 \Cref{fig:depth} shows the performance of different vanilla architectures trained via GD and their associated Nystr\"{o}m approximations. The evaluated networks have the same receptive fields but different depths. In this setting (NTK regime), shallower networks achieve better performance than deeper counterparts. 
 
Interestingly, the fixed CNN filter (via GD) induced by the vanilla architecture performs better than its autoencoder counterpart. Despite having a larger receptive field (i.e., comparing larger patches), we observed that the autoencoder's eigenimages are more blurry than the vanilla CNN.

%In the supplementary, we evaluate the performance of the kernel generated by Adam using the AE architecture: denoising with this fixed kernel obtains an average PSNR of 29.3 dB (see SM I).

Despite being able to denoise with a single noisy image as training data, we note that the evaluated methods are below the performance of color NLM and CBM3D which obtain a PSNR of 30.26 and 31.42 dB respectively. However, \corr{we emphasize that} the goal of this paper is to understand the implicit bias of CNNs rather than provide new state-of-the-art denoisers.

\subsection{Change of weights during training}
\Cref{fig:weight_change} shows the PSNR obtained on the house image, and the change of weights in intermediate layers,  $||\wvec^{t}-\wvec^{0}||_2$ and $\sum_{1<\ell<L} \frac{1}{L-2}\max_{i,j,\alpha}|(w_{i,j,\alpha}^{\ell})^{t}-(w_{i,j,\alpha}^{\ell})^{0}|$ as a function of the number of channels when applying GD and Adam with an autoencoder architecture. For Adam, the denoising performance only improves as the number of channels increases, suggesting that the adaptive kernel property is not due to a finite network effect. As explained in \Cref{sec:adaptive}, when using Adam the weights in intermediate layers suffer a change of $\mathcal{O}(1)$ with respect to the $\ell_2$ norm, hence adapting the filter at initialization, whereas GD attains a change of the order $\mathcal{O}(\ch^{-1/2})$, which corresponds to a fixed filter as $\ch\to\infty$. Furthermore, all individual weights incur a similar small change of order $\mathcal{O}(\ch^{-1/2})$ during training with Adam, suggesting that each weight induces a similar (small) contribution to the network output, in contrast with convolutional sparse coding~\cite{wohlberg2016csc} interpretations, where only a few weights are non-negligible. 

Finally, \Cref{fig:preact} shows the leading eigenvectors of the preactivations of the last hidden layer after 500 iterations of training with GD and Adam. As explained in \Cref{sec:adaptive}, GD does not modify the distribution of the preactivations during training, hence they remain non-informative (low-pass~\cite{cheng2019bdip,Soltanolkotabi2020Denoising}) if noise is placed at the input of the network. However, they carry non-local information when the image is placed at the input. On the other hand, Adam, even with noise at the input, modifies the initial non-informative distribution with non-local features extracted from the target.
%It is worth noting that traditional patch-based methods also apply adaptive filtering to improve their performance, e.g., BM3D recomputes affinities between patches using a pre-filtered image~\cite{dabov2007image}.

\section{Discussion}

\corr{Fully-convolutional trained networks, such as DnCNN~\cite{zhang2017gaussian}, do not perform fully global denoising, as the filter is constrained by the size of the receptive field, which does not cover the full image. There has been recent efforts to construct networks which explicitly exploit non-local self-similarities in a fully global fashion, either via non-local networks~\cite{wang2018non} or using architectures that explicitly operate on noisy patches~\cite{liu2018non, vaksman2020lidia}. 
The setting studied here, i.e., training a network with a single corrupted image, corresponds to global filtering~\cite{talebi2014glide}, as correlations between all patches in the image are considered. Our framework has the potential to combine both training data and the exploitation of self-similarities, e.g., through global filtering and Nystr\"{o}m.}

\section{Conclusions}
We introduced a novel analysis of CNN denoisers trained with a single corrupted image, using the recent discovery of the neural tangent kernel to elucidate the strong links with non-local patch-based filtering methods. As the number of channels of the network tends to infinity, the associated pixel affinity function is available in closed form, thus we can study the properties of the induced filter and \corr{understand the denoising through the NTK's low rank approximation}. These results bring insight about the inductive bias of CNNs in image processing problems: The effective degrees of freedom are significantly smaller than the actual number of weights in the network, being fully characterized by the architecture and initialization of the network.

While the NTK theory accurately predicts the behaviour of networks trained with standard gradient descent, we show that it fails to describe the induced filter when training with the popular Adam optimizer. Interestingly, while Adam and other adaptive gradient optimizers are known to provide worse results than stochastic gradient descent in random features models~\cite{wilson2017marginal}, they play a key role here by adapting the filter with non-local information about the target image in the context of the deep image prior. We believe that understanding better the dynamics and hence the inductive bias of these optimizers, will be a very important step for improving our understanding of CNN models, both for denoising and more general imaging and image analysis problems.
%\section*{Acknowledgements}
%Do you we have to acknowledge some grants?  {-- Our ERC grant, but not for now since the submission needs to be anonymous..}

% \section*{Broader Impact}
% While the paper is essentially theoretical in nature, the authors believe that a better understanding of the inductive bias of convolutional neural networks particularly in the context of imaging and computer vision could have a positive impact in many real-life applications, such as medical imaging or autonomous driving, where mis-estimation can threaten the well-being of people. 
% Hence, improving our understanding of the capabilities and limitations of these models is a fundamental step for a safer usage of CNN-based solutions.

\bibliographystyle{unsrt}  
\bibliography{references} 

\appendix

\section{Assumptions and other observations} \label{sec:assumptions}

\begin{enumerate}
    \item We have omitted the use of biases to simplify the presentation. In the case of relu  non-linearities, the presence of biases would add an additional constant term to the $\Vmap[]$ and $\Vpmap[]$ maps in \cref{eq:vmap,eq:vpmap}~\cite{arora2019cntk}. We also found that the denoising performance did not vary significantly with or without them (for relu non-linearities). Moreover, it has been recently shown that bias-free denoisers generalize better for different noise levels~\cite{Mohan2020Robust}.
    \item We focus on the case where all hidden layers have the same number of channels $\ch$. Our analysis can be easily extended for different number of channels per layer, as long as they all grow at the same rate when taking $\ch\to\infty$~\cite{karakida2019fim}.
    \item Despite we assume that the output $z$ has a single channel for the main derivations, the theory applies to a variable number of channels $\ch_L$, as long as they are significantly smaller than the ones of the hidden layers $\ch$. The extension to multiple channels is provided in \cref{sec:multiple_channels}.
    \item  We drop the dependence of the pre-activations $a^{\ell}$ on the input $x$ to lighten notations. 
    \item For ease of presentation, we focus on the case where all layers have the same image size $d_{\ell}=d$. \Cref{sec:downsampling} extends the results for downsampling and upsampling layers of U-Net and autoencoder architectures.
    %\item Note that a fully-connected network is obtained by setting $d=1$ for all layers, yielding scalar $W_{i,j}$, where an input of size $n$ is defined as a 1-pixel image with $\ch_0=n$ channels. 
    \item It is worth noting that some architectures proposed in the deep image prior paper~\cite{ulyanov2018deep} have a number of input channels of order $\mathcal{O}(\ch)$. However, we noticed that reducing the number of channels does not impact significantly the performance.
    \item To the best of our knowledge, the theory presented here cannot not be straightforwardly applied to networks with batch normalization and max pooling. However, we noted that they do not affect significantly the denoising performance of the networks. 
\end{enumerate}

\section{Forward signal propagation}
\label{sec:forward}
In this section we study the statistics of the signal as it propagates through the neural network. As $\ch\to\infty$, the preactivations at each layer $a_i^{\ell}$ can be well described by a multivariate Gaussian distribution due to the central limit theorem~\cite{neal1995bayesian}. Hence, computing the mean and covariance is enough to fully characterize their distribution. 
For the first hidden layer we have, for each channel $i=1,\dots,\ch$, mean
\begin{align}
   \meanA{1} &= \expected{W_{i,1}^{1}} \expected{x} \\ &= 0
\end{align}
and covariance 
\begin{align}\label{eq:first_layer_cov}
   \SigmaA{1} &= \expected{W_{i,1}^{1}x x^T(W_{i,1}^{1})^T} 
\end{align}
where the independence of weights across different filters was used to simplify the sum. Note that we have dropped the dependence of the mean and covariance on the specific channel $i$, as all channels share the same mean and covariance.  
The expression in \cref{eq:first_layer_cov} consists of pairwise expectations
\begin{align}
 \expected{[W_{i,j}^{\ell}x]_\mu [W_{i,j}^{\ell}x]_v }  = \frac{1}{r^2}\sum_{\mu',v'} x_{\mu'} x_{v'} 
\end{align}
where $\mu'$ and $v'$ are the indices of pixels within patches of size $r \times r$ centered at $\mu$ and $v$ respectively. 
It can be written in a more compact form as 
\begin{equation}
   \SigmaA{\ell} = \Amap[xx^T]
\end{equation}
where the convolution map $\Amap[]: \text{PSD}_n \mapsto \text{PSD}_n$ is defined as~\cite{xiao2018dynamical} 
\begin{equation} \label{eq:A_map}
    [\Amap[\Sigma]]_{\mu,v} =  \frac{1}{r^2}\sum_{\mu',v'} [\Sigma]_{\mu',v'}
\end{equation}
For the following layers we also have zero mean, i.e.,
\begin{align}
   \meanA{\ell} &= \sum_{j=1}^{\ch} \expected{W_{i,j}^{\ell}} \expected{\phi\left(a_{j}^{\ell-1}\right)} \\ &= 0
\end{align}
and a covariance is given by
\begin{align} \label{eq:cnn_cov}
   \SigmaA{\ell} &= \sum_{j=1}^{\ch}  \expected{W_{i,j}^{\ell-1}\act{a_{j}^{\ell-1}}\act{a_{j}^{\ell-1}}^T(W_{i,j}^{\ell-1})^T} 
\end{align}
where the first term of the right hand side is given by 
\begin{align}
 \expected{[W_{i,j}^{\ell}\act{a_{j}^{\ell-1}}]_\mu [W_{i,j}^{\ell}\act{a_{j}^{\ell-1}}]_v }  &= \sum_{\mu',v'} \expected{\act{a_{j,\mu'}^{\ell-1}} \act{a_{j,v'}^{\ell-1}}}
\end{align}
The expression can be written in compact form as 
\begin{equation}
\label{eq:forw_sigma}
   \SigmaA{\ell} = \Amap[\Vmap[\SigmaA{\ell-1}]]
\end{equation}
where
the map $\Vmap[]: \text{PSD}_n \mapsto \text{PSD}_n$ linked to a non-linearity $\act{x}$ is defined as 
\begin{equation}
    \Vmap[\Sigma] = \sigma_w^2\expectedv{\act{h}\act{h^T}}{h\sim \mathcal{N}(0,\Sigma)}
\end{equation}
%As the number of channels of the neural networks goes to infinity $\ch\to\infty$, the variance of the weights should scale as $\sigma_{\wvec^{\ell}}^2\propto \ch^{-1}$ to avoid infinity variance (diverging signal). This strategy is known as \emph{Kaiming He}~\cite{He2015delving} initialization\footnote{There are many similar ones, such as \emph{Glorot} and \emph{LeCun} initialization~\cite{Glorot2010understanding} which differ by a constant value.}.
The $\Vmap[]$-map consists of two-dimensional integrals that are available in closed-form for many activation functions. In the case of relu non-linearities, we have~\cite{cho2009kernel}
\begin{equation}
\label{eq:vmap}
     [\Vmap[\Sigma]]_{\mu,v} = \frac{\sqrt{\Sigma_{\mu,\mu}\Sigma_{v,v}}}{\pi} (\sin(\varphi) + (\pi-\varphi)\cos(\varphi))
\end{equation}
where 
$\varphi=\arccos(\Sigma_{\mu,v}/\sqrt{\Sigma_{\mu,\mu}\Sigma_{v,v}})$. As discussed in \cite{xiao2018dynamical}, $\ell$ repeated applications of the operator given by \cref{eq:vmap} quickly converge to a matrix of the form 
\begin{equation}
\label{eq:v_map_conv}
    [\Sigma]_{\mu,v} = \begin{cases} 
   1 &\mbox{if } \mu=v \\
\kappa_\ell & \mbox{otherwise} \end{cases}
\end{equation}
where $k_\ell$ decreases to zero exponentially fast with depth. Note that the matrix in \cref{eq:v_map_conv} is invariant to the $\Amap[]$ map, as the diagonal elements are averaged with other diagonal elements, whereas the off-diagonal entries are averaged with other off-diagonal ones.

The output $z$ is also characterized by a multivariate Gaussian distribution with
\begin{equation}
   \SigmaZ = \Amap[\Vmap[\SigmaA{L-1}]].
\end{equation}

The main difference between the fully connected and convolutional architectures lies in the covariance $\SigmaA{\ell}$. In the fully connected case, $\Amap[]$ boils down to the identity operator, and $\SigmaA{\ell}$ has an isotropic structure for all layers, whereas the convolutional network presents rich covariances within the pixels of each channel in \cref{eq:cnn_cov}, as $\Amap[]$ cross-correlates different patches of the image.

\subsection{Gaussian process interpretation}
We can use the distribution of an infinite neural network at initialization to define a prior $p(z)=\mathcal{N}(0, \SigmaZ)$ for images, following a Bayesian inference viewpoint~\cite{neal1995bayesian}, a strategy named the Bayesian deep image prior in~\cite{cheng2019bdip}. In the case of standard Gaussian noise $z=y+n$ we have
\begin{align}
    y|z &\sim \mathcal{N}(z,\sigma^2_nI) \\
    z &\sim \mathcal{N}(0, \SigmaZ)
\end{align}
where the posterior distribution is available in closed form
\begin{equation}
    z|y \sim \mathcal{N}\left((I+\sigma^{2}_n\SigmaZ^{-1})^{-1}z, (I\sigma^{-2}_n+\SigmaZ^{-1})^{-1} \right)
\end{equation}
Note that, if iid noise is placed at the input of the network, $\SigmaZ$ does not depend on the noise image $z$ in any way. Moreover, for a relu network, this covariance is given by \cref{eq:v_map_conv}. \Cref{fig:off-diagonal} shows that the off-diagonal elements $\kappa_L$ tend to 1 as the network becomes larger. This prior just promotes constant images.

\section{Backward signal propagation}
\label{sec:backward}
A similar analysis can be made for the propagation of gradients through the network in backwards direction. This is especially useful to study the behaviour of backpropagation training and avoid vanishing or exploding gradients in deep networks. Computing gradients with respect to the weights of the $\ell$th layer can be done using the chain rule:
\begin{equation}
    \der{\loss}{\wvec^{\ell}} =\der{\loss}{z} \der{z}{a^{L-1}}\dots\der{a^{\ell+1}}{a^{\ell}}\der{a^{\ell}}{\wvec^{\ell}}
\end{equation}
We define the gradient as:
\begin{equation}
\label{eq:grad_def}
\grad{\ell}_i\defeq \der{\loss}{z}\der{z}{a^{L-1}}\dots\der{a^{\ell}}{a^{\ell-1}_i}\in \mathbb{R}^{d} 
\end{equation} 
with $\grad{L}\defeq\der{\loss}{z}$. % then 
%\begin{align}
%    \der{\loss}{\wvec^{\ell}_{i,j}} & =\grad{\ell}_{i}\der{a^{\ell}}{\wvec^{\ell}_{i,j}} \\
%    &=  \grad{\ell}_i \actd{a_{j}^{\ell}}^T.
%\end{align}
For a squared loss, the gradient at the last layer  is
\begin{equation}
    \grad{L} = z - y.
\end{equation}
Assuming that independence between gradients and preactivations~\cite{xiao2018dynamical}\footnote{This assumption is formally justified in a recent work~\cite{yang2019mean}.}, we have for each channel $i=1,\dots,\ch$ of  layer $L-1$ 
\begin{equation}
    \grad{L-1}_i =  \diag{\actd{a_{i}^{L-1}}}(W_{1,i}^{L})^T \grad{L}
\end{equation}
which has zero mean and covariance given by
\begin{equation}
\label{eq:sigmaD L-1}
    \SigmaD{L-1} = \frac{1}{\ch} \Vpmap[\SigmaA{L-1}]\circ \Amap[\SigmaD{L}]
\end{equation}
where the map $\Vpmap[]: \text{PSD}_n \mapsto \text{PSD}_n$ is defined as 
\begin{equation}\label{eq:vpmap}
     \Vpmap[\Sigma] = \sigma_w^2\expectedv{\actd{h}\actd{h^T}}{h\sim \mathcal{N}(0,\Sigma)}
\end{equation}
The expected values are available in closed form for many non-linearities.
We can use the following recursive formula to compute the rest of the layers $\ell=L-2,\dots,1$
\begin{equation}
    \grad{\ell}_i =  \sum_{j=1}^{C^{\ell}} \diag{\actd{a_{i}^{\ell-1}}}(W_{j,i}^{\ell})^T \grad{\ell+1}_j
\end{equation}
Computing the propagation recursively in backwards direction, we have $ \mu_{\grad{\ell}} =0$ and covariance 
\begin{align}
\label{eq:sigmaD recursive}
    \Sigma_{\grad{\ell}_i}  &= % \sum_{j=1}^{C^{\ell}}\expected{\diag{\actd{a^{\ell-1}_i}} (W_{j,i}^{\ell})^T  \grad{\ell+1}_j (\grad{\ell+1}_j)^T W_{j,i}^{\ell} \diag{\actd{a_i^{\ell-1}}} } \\
   \Amap[\SigmaD{\ell+1}] \circ \Vpmap[\SigmaA{\ell}]
\end{align}
For relu non-linearities the $\Vpmap[]$ map is computed as 
\begin{equation}
     [\Vpmap[\Sigma]]_{\mu,v} = 1-\frac{1}{\pi}\arccos \frac{\Sigma_{\mu,v}}{\sqrt{\Sigma_{\mu,\mu}\Sigma_{v,v}}}
\end{equation}
which as with the $\Vmap[]$ counterpart\footnote{Note that the discontinuity of the relu function at 0 is unimportant here due to the expectation operator.}, repeated applications of this map converge exponentially fast to the simple matrix structure in \cref{eq:v_map_conv}.

\section{Neural Tangent Kernel} \label{sec:ntk}
\begin{figure}
  \centering
    \includegraphics[width=.4\textwidth]{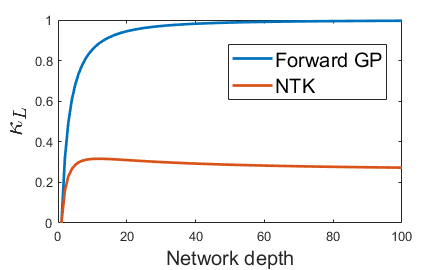}
  \caption{Off-diagonal elements of the filtering matrix associated with the Gaussian process at initialization and the neural tangent kernel with iid noise input.}
  \label{fig:off-diagonal}
\end{figure}

In this section, we will denote all the trainable network parameters at iteration $t$ as $\wvec^{t}$. Consider training a network via gradient descent\footnote{A very similar analysis can be done for gradient flow and stochastic gradient descent~\cite{lee2019wide}}, that is 
% \begin{align}
%     \der{\wvec^{\ell}}{t} = -\lr \der{\loss}{\wvec^{\ell}} 
% \end{align}
\begin{align}
\label{eq:weight_update}
    \wvec^{t+1} = \wvec^{t} -\lr \der{\loss}{\wvec}(\wvec^{t})
\end{align}
We can study the evolution of the function defined by the weights $z^{t}\defeq z(\wvec^{t})$, using a first order Taylor expansion, i.e.,
\begin{align}
\label{eq:output_taylor}
 z^{t+1} &\approx  z(\wvec^{t}) + \der{z}{\wvec} (\wvec^{t+1}-\wvec^{t}) \\
 &\approx z^{t} - \lr \der{z}{\wvec} \der{\loss}{\wvec} \\
 &\approx z^{t} - \lr \der{z}{\wvec} (\der{z}{\wvec})^T \der{\loss}{z}
\end{align}
where we have used \cref{eq:weight_update} in the second line and the chain rule in the third line. The neural tangent kernel (NTK) is given by
\begin{align}
\label{eq:ntk}
    \NTK_{L} &= \der{z}{\wvec} (\der{z}{\wvec})^T \\
    &=\sum_{\ell,i,j,\alpha}   \der{z}{w_{i,j,\alpha}^{\ell}} (\der{z}{w_{i,j,\alpha}^{\ell}})^T 
\end{align}
We can start with the base case,
\begin{align}
    \NTK_{2} = c\Vmap[\Amap[x x^T]]
\end{align}
and notice the following recursive formulation
\begin{align}
\label{eq:NTK_recursive_form}
    \NTK_{\ell} &=  \der{a^\ell_i}{\wvec^{\ell}}(\der{a^\ell_i}{\wvec^{\ell}})^T  + \der{a^\ell_i}{a^{\ell-1}} \NTK_{\ell-1} (\der{a^\ell_i}{a^{\ell-1}})^T \\
    &= \sum_{j=1}^{\ch}  \Amap[\act{a_j^{\ell-1}}\act{a_j^{\ell-1}}^T] + W_{i,j}^{\ell}\diag{\actd{a_j^{\ell-1}}}\NTK_{\ell-1}\diag{\actd{a_j^{\ell-1}}}(W_{i,j}^{\ell})^T
\end{align}
where $\wvec^{\ell}$ denotes the weights corresponding to layer $\ell$. The learning rate $\eta$ is chosen of order $\mathcal{O}(\ch^{-1})$, in order to converge to global minimum~\cite{karakida2019fim}. Without loss of generality, we use $\eta=\gamma\ch^{-1}$ for the following derivations, where $\gamma$ is $\mathcal{O}(1)$ and chosen such that the neural tangent kernel has its eigenvalues bounded by 1.
As shown in~\cite{yang2019mean}, for an infinite number of channels $\ch\to\infty$, due to the law of large numbers we have
\begin{align}
    \lr\NTK_{\ell} = \SigmaA{\ell} + \Amap[\Vpmap[\SigmaA{\ell}] \circ\lr\NTK_{\ell-1}]
\end{align}
which is a fixed (deterministic) matrix. %Note that for a single input and a fully connected structure the resulting NTK is diagonal. 
As a function of the input image (or noise) patches, the NTK defines a kernel acting on pairs of input patches $x_1$ and $x_2$, i.e., $k(x_1,x_2):\mathbb{R}^{d_0}\times\mathbb{R}^{d_0}\mapsto \mathbb{R}_+$. As discussed in the main paper, if iid noise is placed at the input, the resulting Gram matrix is given by \cref{eq:v_map_conv} with $\kappa_L$ as shown in \Cref{fig:off-diagonal}.

For a squared loss $\loss=\frac{1}{2}||z-y||_2^2$, the dynamics of \cref{eq:output_taylor} can be written as
\begin{align}
    \label{eq:output_flow}
    z^{t+1} &= z^{t} + \lr\NTK_{L} \left(y-z^t \right) \\
    &= (I-\lr\NTK)^{t+1} z^{0} +\sum_{k=1}^{t} (\lr\NTK_{L})^{k} y
\end{align}
with initial condition $z^{0}$ given by the Gaussian process initialization described in \Cref{sec:forward}. The expression for $z^{t}$ can be simplified further by noting that the learning rate has to be chosen such that $\lr\NTK$ has its eigenvalues bounded from above by 1 (to avoid a diverging gradient descent). Hence, as $I-\lr\NTK$ is invertible, we can apply the geometric series formula  
\begin{equation}
\label{eq:output_dynamics}
    z^{t} = (I-\lr\NTK_{L})^{t} z^{0} + (I-\lr\NTK_{L})^{-1} \left(I-(\lr\NTK_{L})^{t} \right) y
\end{equation}
Note that the only random component of this equation is the Gaussian process initialization $z^{0}$. As $z^{t}$ is an affine transformation of a Gaussian process, it is also itself a Gaussian process for every iteration $t$. Hence, we have 
\begin{equation}
 z^{t} \sim \mathcal{N}\left((I-\lr\NTK)^{-1} \left(I-(\lr\NTK)^{t} \right) y, (I-\lr\NTK_{L})^{t}\Sigma_{z}(I-\lr\NTK)^{t}\right)
\end{equation}
It is easy to see that $z^{t}$ converges at an exponential rate towards a singular distribution centered at $y$ as $t\to\infty$. 

\section{Multiple input and output channels} \label{sec:multiple_channels}
The theory applies for any number of input and output channels, as long as they are much smaller than the number of hidden channels $\ch$. A multi-channel input modifies the computation in the first layer \cref{eq:first_layer_cov}. In this case,  first multiplying the patches channel-wise and then summing the result, that is
\begin{align}
   \SigmaA{1} &= \sum_{j=0}^{\ch_0} \expected{W_{i,j}^{1}x_j x_j^T(W_{i,j}^{1})^T} 
\end{align}
where $x_j$ denotes the $j$th channel of the input, and the corresponding infinite-width operator is computed as 
\begin{align}
   \SigmaA{1} &= \frac{1}{\ch_0}\sum_{j=0}^{\ch_0} \Amap[x_j x_j^T]
\end{align}
Hence, the pixel affinity function is now defined for a receptive field $d_0\leq d$, and patches $x_1$ and $x_2$ of $\ch_0$ channels as 
\begin{equation}
    k(x_1,x_2):\mathbb{R}^{\ch_0d_0}\times\mathbb{R}^{\ch_0d_0}\mapsto \mathbb{R}_+.
\end{equation}

Multiple output channels are computed separately using the same filtering matrix, i.e.,
\begin{equation}
    z_i^{t+1} = z_i^{t} + \eta\NTK_L (y-z_i^{t})
\end{equation}
for $i=1,\dots, \ch_L$. Note that both the color versions of NLM and BM3D do a similar procedure, computing the filtering matrix with luminance (i.e., a linear combination of the RGB channels), and apply the filtering process to each channel separately.

\section{Downsampling and upsampling layers} \label{sec:downsampling}

Downsampling can be achieved either via 2-strided convolutional layers or directly with linear downsampling operations, such as bilinear or nearest neighbor downsampling. Strided convolutions are a straightforward extension of the $\Amap[]$ operator defined in \cref{eq:A_map}, summing over strided patches instead of contiguous ones. Linear downsampling operations can be expressed as a matrix vector product applied channel-wise, i.e., $a^{\ell+1}_i = Da^{\ell}_i$ where $D\in\mathbb{R}^{d\times d/2}$ is a fixed matrix given by downsampler (bilinear, nearest neighbor, etc.). The covariance of $a^{\ell+1}_i$ is then
\begin{equation}
    \SigmaA{\ell+1} = D\SigmaA{\ell} D^{T}.
\end{equation}

Upsampling is generally performed with bilinear or nearest neighbor layers, as transposed convolutions provide worse results~\cite{ulyanov2018deep}. These are analogous to the downsampling case, but with an upsampling matrix $U\in\mathbb{R}^{d/2\times d}$, that is
\begin{equation}
    \SigmaA{\ell+1} = U\SigmaA{\ell} U^{T}.
\end{equation}

\section{Nystr\"{o}m denoising} 
\label{sec:nystrom}
The Nystrom method approximates the first $m$ eigenvectors of the NTK matrix by computing only a subset of $m\ll d$ columns~\cite{seeger2001nystrom}, i.e., the sub-matrix
\begin{equation}
\NTK_{d,m} = \begin{bmatrix}
\NTK_{m,m} \\
\NTK_{d-m,m}  
\end{bmatrix}    
\end{equation}
We first perform a singular value decomposition of the small sub-matrix $\NTK_{m,m}=\sum_{i=1}^{m}\tilde{\lambda}_i \tilde{v}_i \tilde{v}_i^{T}$, and then approximate the eigenvectors and eigenvalues of the full matrix as 

\begin{align}
    v_i &= \sqrt{\frac{m}{d}}\frac{1}{\tilde{\lambda}_i} \NTK_{d,m} \tilde{v}_i  \\
    \lambda_i &= \frac{d}{m}\tilde{\lambda}_i 
\end{align}
We fix $m=0.02d$, which allows us to compute most of the $md$ pixel affinities in parallel on the GPU. The selection of columns is done similarly to global image denoising~\cite{talebi2014glide}, choosing a random selection of pixels uniformly distributed in space. Before applying the denoising procedure, we scale the eigenvalues, such that the maximum eigenvalue is 1.

\section{Architectures} \label{sec:architectures}

\subsection{Vanilla CNN}
\Cref{tab:vanilla} shows the configuration used for the vanilla CNN results with $\ch=512$ channels per hidden layer. The network has a total of 187,392 trainable weights. 

\begin{table}
\centering
\begin{tabular}{|c|c|c|}
\hline
Module & Function & Infinite-channel forward operator \\ \hline
input & 3 channel RGB image &  \\ \hline
conv1 & $11\times 11$ pixel convolution & $\Amap[]$ with $r=11$ \\ \hline
relu1 & relu activation $\max(x, 0)$ & $\Vmap[]$ \\ \hline
conv2 & $1\times 1$ pixel convolution & $\Amap[]$ with $r=1$ \\ \hline
output & 3 channel RGB image &  \\ \hline
\end{tabular}
\caption{Vanilla configuration with a single-hidden layer.}
\label{tab:vanilla}
\end{table}

\subsection{Autoencoder}
\Cref{tab:unet} shows the configuration used for the autoencoder results with $\ch=128$ channels per hidden layer. The network has a total of 1,036,032 trainable weights.

\subsection{U-Net}
The U-Net considered in this paper shares the same architecture and number of weights than the autoencoder, adding skip connections at each level. 

\begin{table}
\centering
\begin{tabular}{|c|c|c|}
\hline
Module & Function & Infinite-channel forward operator \\ \hline
input & 3 channel RGB image &  \\ \hline
convd1 & $3\times 3$ convolution & $\Amap[]$ with $r=3$ \\ \hline
relu1 & relu activation $\max(x, 0)$ & $\Vmap[]$ \\ \hline
down1 & Bilinear downsampling  & $D$ \\ \hline
convd2 & $3\times 3$ convolution & $\Amap[]$ with $r=3$ \\ \hline
relu2 & relu activation $\max(x, 0)$ & $\Vmap[]$ \\ \hline
down2 & Bilinear downsampling  & $D$ \\ \hline
convd3 & $3\times 3$ convolution & $\Amap[]$ with $r=3$ \\ \hline
relu3 & relu activation $\max(x, 0)$ & $\Vmap[]$ \\ \hline
down3 & Bilinear downsampling  & $D$ \\ \hline
convd4 & $3\times 3$ convolution & $\Amap[]$ with $r=3$ \\ \hline
relu4 & relu activation $\max(x, 0)$ & $\Vmap[]$ \\ \hline
conv4 & $3\times 3$ convolution & $\Amap[]$ with $r=3$ \\ \hline
up1 & Bilinear upsampling  & $U$ \\ \hline
convu1 & $3\times 3$ convolution & $\Amap[]$ with $r=3$ \\ \hline
relu5 & relu activation $\max(x, 0)$ & $\Vmap[]$ \\ \hline
up2 & Bilinear upsampling  & $U$ \\ \hline
convu2 & $3\times 3$ convolution & $\Amap[]$ with $r=3$ \\ \hline
relu6 & relu activation $\max(x, 0)$ & $\Vmap[]$ \\ \hline
up3 & Bilinear upsampling  & $U$ \\ \hline
convu3 & $3\times 3$ convolution & $\Amap[]$ with $r=3$ \\ \hline
relu7 & relu activation $\max(x, 0)$ & $\Vmap[]$ \\ \hline
convu4 & $1\times 1$ convolution & $\Amap[]$ with $r=1$ \\ \hline
output & 3 channel RGB image &  \\ \hline
\end{tabular}
\caption{Autoencoder configuration with bilinear downsampling and upsampling layers.}
\label{tab:unet}
\end{table}

\section{Additional results} \label{sec:additional_exp}
In all the denoising experiments, we normalize the corrupted images by subtracting 0.5 from all pixels, such that they defined in the centered interval $[-0.5,5]$. Before computing the PSNR, we denormalize the images by summing 0.5 to all pixels and clipping, such that all pixels are in the interval $[0, 1]$.

\subsection{Denoising examples}
\begin{figure}
  \centering
    \includegraphics[width=1\textwidth]{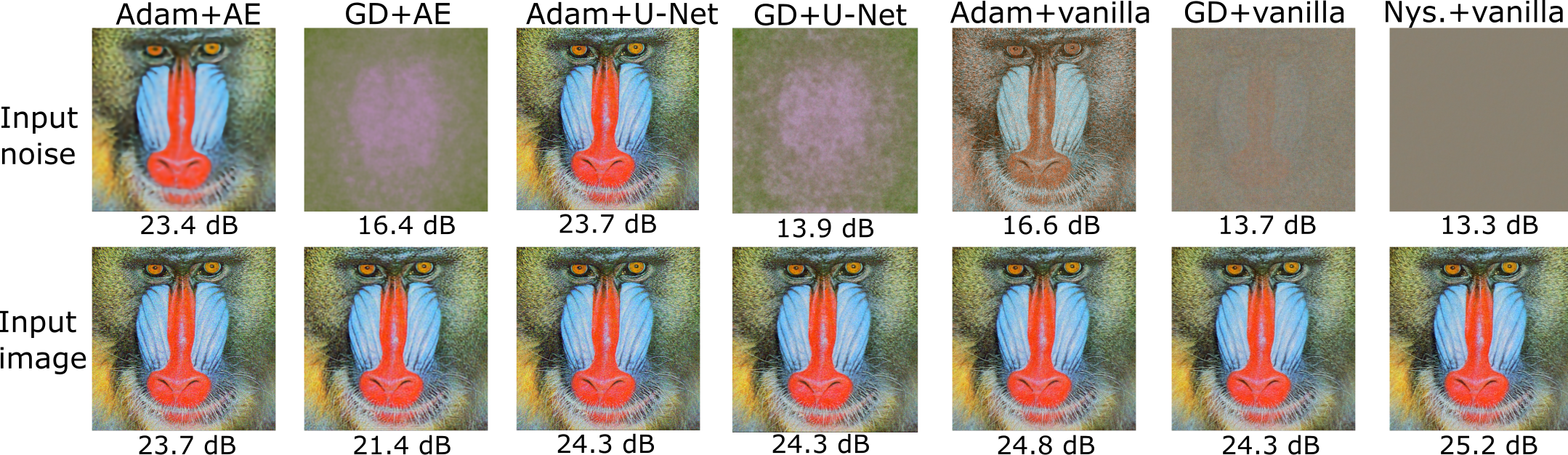}
  \caption{Results for the `baboon' image. PSNR values are reported below each restored image. The best results are obtained by the Nystr\"{o}m approximation of a vanilla CNN filter.}
  \label{fig:baboon comparison}
\end{figure}

\begin{figure}
  \centering
    \includegraphics[width=1\textwidth]{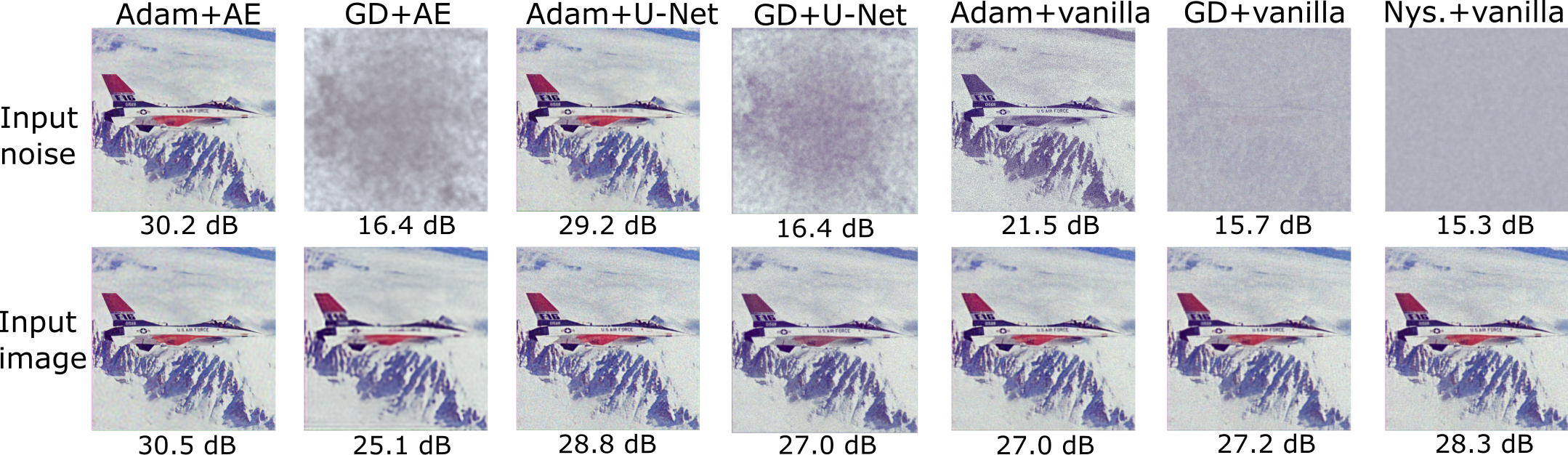}
  \caption{Results for the `F16' image. PSNR values are reported below each restored image. The best results are obtained by an autoencoder trained with Adam, which is able to provide smoother estimates while preserving sharp edges.}
  \label{fig:f16 comparison}
\end{figure}

The deep image prior setting (autoencoder, noise input and Adam optimizer), performs very well in images with large piece-wise smooth patches, such as the `house' image shown in the main paper or the `F16' image in~\Cref{fig:f16 comparison}, but does not provide good reconstructions in images with noise-like textures, such as the `baboon' shown in \Cref{fig:baboon comparison}. The best performing denoiser for this image is the closed form filter associated with a vanilla CNN, approximated with Nystr\"{o}m. 

\subsection{Additional noise levels}
We evaluate the best-performing denoisers (autoencoder with noise or image input trained using Adam and Nystr\"{o}m approximation of a vanilla CNN) for iid Gaussian noise with standard deviations of $\sigma=5$ (low noise) and $\sigma=100$ (high noise). \Cref{tab:noise levels} shows the results for the dataset of 9 color images~\cite{dabov2007image}. Inputting the image when using Adam achieves an improvement of 1.8 dB in the low-noise case, whereas it provides slightly worse (0.3 dB) results in the high noise case.

\begin{table}
\centering
\begin{tabular}{|c|c|c|c|}
\hline
 & AE/Adam/noise & AE/Adam/image & Vanilla/Nystr\"{o}m/image \\ \hline
$\sigma=5$ &33.5 & \textbf{35.3} & 34.5  \\ \hline
$\sigma=100$ &  \textbf{24.4} & 24.1 & 22.3 \\ \hline
\end{tabular}
\caption{Average PSNR [dB] obtained by the best-performing algorithms for different noise levels.}
\label{tab:noise levels}
\end{table}

\subsection{Epoch count}
\Cref{tab:epoch_results} shows the average epoch-count of all methods for the 9 color image dataset. Inputting the image instead of noise reduces the number of iterations when optimizing with Adam, as the induced filtering matrix is better conditioned. Gradient descent requires many more iterations than Adam as it does not uses any momentum. As discussed in the main paper, the filtering matrix associated with a vanilla CNN and noise input is so ill-conditioned that gradient descent does not converge even after $10^{6}$ iterations. 

\begin{table*}[!h]
\centering
\begin{tabular}{l|c|c|c|c|c|c|}
\cline{2-7}
 & \multicolumn{2}{c|}{Vanilla CNN} & \multicolumn{2}{c|}{U-Net}  & \multicolumn{2}{c|}{Autoencoder} \\ 
\cline{2-7}
 & Noise & Image & Noise & Image & Noise & Image \\ \hline
\multicolumn{1}{|l|}{Adam} & 145340 & 64 & 7692 & 74 & 10248 & 5088\\ \hline
\multicolumn{1}{|l|}{Gradient descent} &  $>10^{6}$ & 69526 & 50054 & 5506 & 50355 & 286042 \\ \hline
\multicolumn{1}{|l|}{Nystr\"{o}m} & 368 & 504  \\ \cline{1-3}
\end{tabular}
\caption{Average epoch-count by different combinations of network architecture, input and optimizer on the dataset of 9 color images~\cite{dabov2007image}.}
\label{tab:epoch_results}
\end{table*}

\end{document}